\pdfoutput=1

\documentclass[11pt]{article}

\usepackage{ACL2023}
\usepackage{graphicx}
\usepackage{todonotes}

\usepackage{times}
\usepackage{latexsym}

\usepackage[T1]{fontenc}

\newcommand{\fig}[3][0.9]{
\begin{figure}
\centering
\includegraphics[width=#1\columnwidth]{figure/#2}
        \caption{#3}~\label{fig:#2}
\end{figure}}

\usepackage[utf8]{inputenc}

\usepackage{microtype}

\usepackage{inconsolata}

\title{A LLM Benchmark based on the Minecraft Builder Dialog Agent Task}

  \author{Chris Madge\\
  Queen Mary University of London \\
  \texttt{c.j.madge@qmul.ac.uk} \\\And
  Massimo Poesio\\
  Queen Mary University of London \\
  \texttt{m.poesio@qmul.ac.uk} \\}

\begin{document}
\maketitle
\begin{abstract}
In this work we proposing adapting the Minecraft builder task into an LLM benchmark suitable for evaluating LLM ability in spatially orientated tasks, and informing builder agent design.  Previous works have proposed corpora with varying complex structures, and human written instructions.  We instead attempt to provide a comprehensive synthetic benchmark for testing builder agents over a series of distinct tasks that comprise of common building operations.
We believe this approach allows us to probe specific strengths and weaknesses of different agents, and test the ability of LLMs in the challenging area of spatial reasoning and vector based math.
\end{abstract}

\section{Introduction}
The development of conversational agents able to operate in virtual world environments  has  long been of interest in AI \cite{winograd1972understanding}.
In recent years, much of this research has focused on developing agents able to operate in game environments. 
Game environments
provide an ideal sandbox for studying task-oriented
conversational agents in games \cite{szlam2019build},
which has motivated the development of multiple platforms in which such research can be carried out  \cite{johnson2016malmo,urbanek2019light,callison2022dungeons}
\cite{gray2019craftassist,ogawa2020gamification,kohn2020mc}, data gathering exercises \cite{narayan2019collaborative,jayannavar2020learning,mohanty2022collecting} and competitions \cite{iglucomp}.

The goal of this work is to propose a synthetic benchmark like dataset for testing LLMs on text-based spatial reasoning and vector based math.  Existing work has designed a series of benchmarks to test how LLMs perform on tasks that are outside the scope of ordinary token prediction \cite{srivastava2022beyond,wu2023smartplay}. However, to our knowledge, the requirement for spatial reasoning is uncommon, and does not feature the requirement for 3D construction.   Prior to LLM benchmarking, other tasks have been proposed for testing text-based spatial reasoning however, these are unlikely to motivate the combined vector mathematics, disambiguation or structure required by this task \cite{weston2015towards,shi2022stepgame,mirzaee2022transfer}.

Our particular benchmark is inspired by the virtual world environment ``Minecraft Builder Task'' proposed in \cite{jayannavar2020learning}, in which, given text based instructions from an architect, a builder must take actions to complete a structure, without being able to see the target structure. Previous work has looked at using LLMs in this setting \cite{madge2024large,kranti2024retrievalaugmentedcodegenerationsituated}, and while the performance looks promising, spatial reasoning and vector mathematics remain a challenging task for LLMs \cite{bang2023multitask}.

Aside from being an interesting benchmark of ever evolving LLM ability outside text-based tasks, we hope this may also inform builder agent designers on specific strengths and weaknesses of their approach. Looking through the datasets we have identified some common patterns that occur and produced scenarios to test against those.

Beyond proposing this benchmark, we provide some early discussions over our experience on testing them with \emph{Llama-3-70b-Instruct}, our methods of addressing those challenges, and an evaluation of those methods.

\section{Our Approach}

Previous corpora have shapes that typically represent objects. However, it would appear that the final description of the object the structure represents has little utility in communicating the desired structure.   We identify common patterns used to deliver instructions, and take a rule driven approach to produce architect instructions for the builder around varied set of arrangements of blocks within the context of those patterns.

To validate our benchmark, we test it against a few different prompting approaches.  We take a zero shot approach, a few shot approach, and finally, Chain of Thought \cite{wei2022chain}.

As we further describe our approach in this section, we motivate it through existing examples taken from a previous corpus \cite{narayan2019collaborative}. Naturally, there are multiple ways of representing an object in voxel form, and as the representation is somewhat abstract, given that it is in voxel form, it may not be evident to both parties what object the structure is intended to represent  (e.g. \ref{sec:appendix-1}).
When the final label is used, it tends to be used by the builder to verify the architects instructions in the conclusion of the conversation, rather than by the architect as part of the instruction (e.g. \ref{sec:appendix-2}). 
When the structure is likened to an object, it is almost always accompanied by specific block by  block instructions, and not in isolation (e.g. \ref{sec:appendix-3}).

We find more commonly, the instructions take one of three forms, that we discuss in the following subsections.

\subsection{Absolute Addressing}
At the beginning of the dialog for a task, or when creating a new separated substructure, an architect will need to refer to a space in the grid without the use of an existing reference point, so the references are given to the extent of the grid itself, e.g. \ref{sec:appendix-4}.
We refer to this as absolute addressing.  To benchmark this ability, we produce a test in which the agent is challenged to place a block in every single position in the grid on the first three levels.


\subsection{Relative Addressing}
Relative addresses are possibly the most common type, given throughout the dialog in reference to existing block positions (e.g. \ref{sec:appendix-5}).
To test this, we require the builder place a block in every direction adjacent to an existing block (as shown in Figure \ref{fig:rel}).  Three other blocks are always present in different colours to serve as distractors.   We repeat this test with removal, instead of addition.

\fig[1]{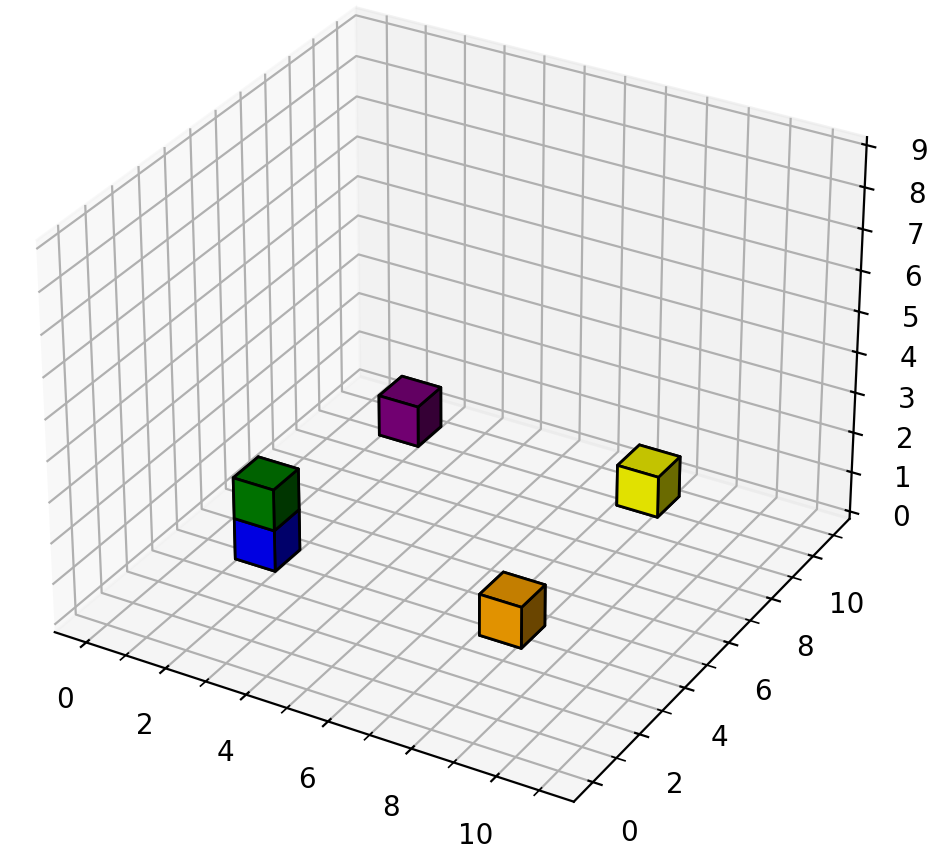}{Relative positioning task, placing a green block on top of an existing blue block}

\subsection{Primitive Shapes}
When commands to build structures comprising of multiple blocks are given, they are typically primitive shapes, such as rows of blocks, or towers, e.g. \ref{sec:appendix-6}.
We test four separate primitives, a row, a tower/stack, a cube and a rectangle.

\section{Results}

Table \ref{tbl:results} shows a range of scores between approaches, representing what might be expected from applying the different prompting techniques.

\begin{table}
\centering
\begin{tabular}{r|c|c}
& Zero Shot &  CoT \\
\hline
Absolute Addressing & 42.98 &  \textbf{76.5}  \\
Relative Addressing & 82.02 &  \textbf{95.8} \\
Primitive Shapes & 59.02 &  \textbf{60.3} \\
\end{tabular}
\caption{Results}
\label{tbl:results}
\end{table}


We believe this methodology may be useful in discovering the weak points in agents, and informing the method of addressing them.  For example, one of the main points identified, is without the Chain of Thought approach, the LLM often neglects to compute one of the axis. In addition, despite the LLM being instructed to apply the right handed 3d coordinate convention, were $Z$ positive for south, south is frequently associated with negative (left handed).  This can be avoided by reinforcing this notion through a few shot example. 

\section{Conclusion}

In this work we propose a new LLM benchmark based around a Minecraft-like task.  We test the validity of this benchmark by applying a few basic strategies to see how this challenges a current LLM.

\section*{Acknowledgements}

This research was funded by ARCIDUCA, EPSRC EP/W001632/1

\bibliography{acl2023}
\bibliographystyle{acl_natbib}

\appendix

\section{Appendix}
\subsection{\texttt{B1-A3-C8-1522432497234}}
\label{sec:appendix-1}
\begin{quote}
\begin{description}\itemsep0pt
\item[Builder] its a table?
\item[Architect] i don't know what it is
\end{description}
\end{quote}

\subsection{\texttt{B1-A3-C4-1522432009099}}
\label{sec:appendix-2}
\begin{quote}
\begin{description}\itemsep0pt
\item[Builder] its a flower!
\item[Architect] yes it is, you are very observant builder
\end{description}
\end{quote}

\subsection{\texttt{B1-A3-C1-1522435497386}}
\label{sec:appendix-3}
\begin{quote}
\begin{description}
\item[Architect] now we must create the bell. please start by extending 4 orange blocks down from the middle purple block, as if it were hanging
\end{description}
\end{quote}

\subsection{\texttt{B3-A2-C12-1522445699382}}
\label{sec:appendix-4}
\begin{quote}
\begin{description}
\item[Architect] In the upper left corner place a purple block
\end{description}
\end{quote}

\subsection{\texttt{B3-A2-C23-1522447244858}}
\label{sec:appendix-5}
\begin{quote}
\begin{description}
\item[Architect] add another green block below each red one you added
\end{description}
\end{quote}

\subsection{\texttt{B1-A3-C3-1522431780184}}
\label{sec:appendix-6}
\begin{description}
\item[Architect] build a 2x1 structure that is blue
\end{description}

\end{document}